\documentclass[10pt,journal,compsoc]{IEEEtran}
\usepackage{graphicx}
\usepackage{subfigure}
\usepackage{threeparttable}
\usepackage{soul, color, xcolor}

\soulregister\cite7
\soulregister\ref7

\ifCLASSOPTIONcompsoc
  \usepackage[nocompress]{cite}
\else
  \usepackage{cite}
\fi
\ifCLASSINFOpdf
\else
\fi
\begin{document}
%
\title{Spreeze: High-Throughput Parallel Reinforcement Learning Framework}
%
%
%
%

\author{Jing~Hou,
        Guang~Chen*,~\IEEEmembership{Member,~IEEE},
        Ruiqi~Zhang,
        Zhijun~Li,~\IEEEmembership{Fellow,~IEEE},
        Shangding~Gu,
        and~Changjun~Jiang   
\IEEEcompsocitemizethanks{
\IEEEcompsocthanksitem
 *G. Chen is the corresponding author.  

\,\,E-mail address: guangchen@tongji.edu.cn
}
}

%
%

\markboth{Preprint Version}
{Shell \MakeLowercase{\textit{et al.}}: Bare Demo of IEEEtran.cls for Computer Society Journals}
%



\IEEEtitleabstractindextext{%
\begin{abstract}
The promotion of large-scale applications of reinforcement learning (RL) requires efficient training computation. While existing parallel RL frameworks encompass a variety of RL algorithms and parallelization techniques, the excessively burdensome communication frameworks hinder the attainment of the hardware's limit for final throughput and training effects on a single desktop.
In this paper, we propose Spreeze, a lightweight parallel framework for RL that efficiently utilizes a single desktop hardware resource to approach the throughput limit. We asynchronously parallelize the experience sampling, network update, performance evaluation, and visualization operations, and employ multiple efficient data transmission techniques to transfer various types of data between processes. The framework can automatically adjust the parallelization hyperparameters based on the computing ability of the hardware device in order to perform efficient large-batch updates. Based on the characteristics of the ``Actor-Critic'' RL algorithm, our framework uses dual GPUs to independently update the network of actors and critics in order to further improve throughput.
Simulation results show that our framework can achieve up to 15,000Hz experience sampling and 370,000Hz network update frame rate using only a personal desktop computer, which is an order of magnitude higher than other mainstream parallel RL frameworks, resulting in a 73\% reduction of training time. Our work on fully utilizing the hardware resources of a single desktop computer is fundamental to enabling efficient large-scale distributed RL training.

\end{abstract}

\begin{IEEEkeywords}
Reinforcement learning, framework, asynchrony, shared memory, model parallelism.
\end{IEEEkeywords}}

\maketitle

\IEEEdisplaynontitleabstractindextext

%
\IEEEpeerreviewmaketitle

\IEEEraisesectionheading{
\section{Introduction}
\label{sec:introduction}}

The recent triumphs in reinforcement learning (RL) owe much to the community's relentless pursuit of rendering RL training faster and more efficient \cite{liang2018rllib, hoffman2020acme}.
In contrast to supervised learning, RL trained in simulated environments necessitates constant time-consuming environment exploration during agent updates to gather experience.
As RL is applied to solve increasingly challenging tasks with larger state and action spaces, the computational time complexity of standard model-free RL becomes $O(S^2 A)$, leading to a dramatic increase in training time.
It is crucial to highlight that RL is pivotal in various real-world applications, such as robotics, finance, and healthcare, where decision-making in complex and dynamic environments is a central challenge. The ability of RL to adapt and learn from experience positions it as a fundamental technology for addressing intricate problems that traditional algorithms struggle to solve.
Furthermore, due to the randomness of the interactive generation and collection of experience in RL, RL training is not as stable as supervised learning training. Researchers are required to conduct repeated experiments many times to validate the performance of their algorithms \cite{agarwal2021deep}.

For tasks of heightened complexity, extensive training times impede the swift development of RL techniques. The computational demand necessitates a shift towards faster RL training, motivating the quest for efficient methodologies. The need is not only for speed but also for stability, as prolonged training times exacerbate RL's inherent instability, hindering advancements in algorithmic performance.
Efficient RL training is not only a matter of practicality but a key enabler for unlocking the potential of RL in addressing complex real-world problems. The significance extends beyond mere computational efficiency to the ability to tackle and conquer challenges that were previously deemed insurmountable.
To expedite training, formidable server setups equipped with numerous CPU and GPU cores and ample memory are employed. However, a poorly designed RL framework risks underutilizing these computing resources, emphasizing the critical role of an effective parallel framework. The motivation is not merely speed but the optimized utilization of powerful computing devices.

Constrained by many factors, it is challenging to significantly increase the base clock frequency of CPU and GPU \cite{komoda2013power}.
Therefore, the core of algorithm acceleration is to utilize multi-core CPU and GPU for parallelization \cite{abughalieh2019survey}.
Supervised learning lends itself well to GPU parallelization acceleration because the data is prepared in advance and the data order is not considered.
However, RL algorithms need to constantly interact with the environment to obtain new data, so they are more complex to be parallelized compared to supervised learning algorithms.

\begin{figure*}[t]
	\centering
    \includegraphics[width=1.0\linewidth]{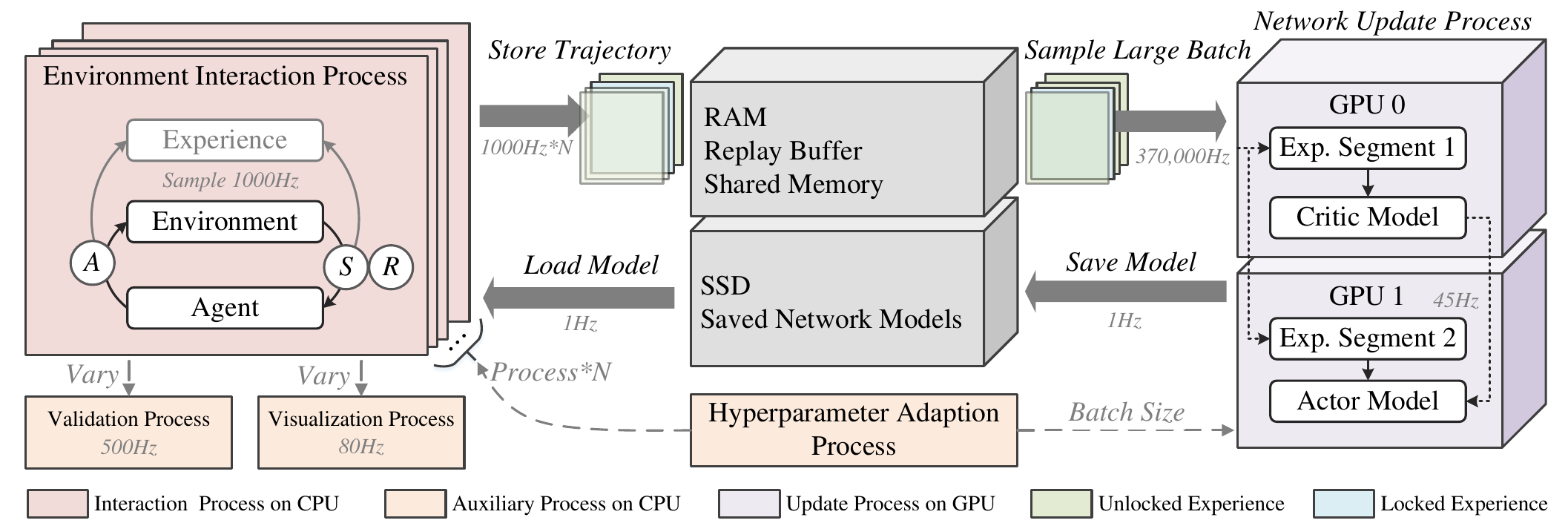}
	\caption[]{\textbf{Overview of our Spreeze architecture.} The experience sampling process on the CPU and the network update process on the GPU transfer experience through shared random-access memory (RAM) and synchronize the network through solid-state drive (SSD). The shown throughput takes the pybullet Walker-2d task as an example.}
	\label{fig_spreeze}
\end{figure*}

The RL community has actively pursued faster and more efficient training methods, including parallel algorithms like A3C \cite{mnih2016asynchronous}, APE-X \cite{horgan2018distributed}, and IMPALA \cite{espeholt2018impala}, as well as general algorithm frameworks such as RLlib \cite{liang2018rllib}, Acme \cite{hoffman2020acme}, and rlpyt \cite{stooke2019rlpyt}, leveraging GPU parallelization for high-speed data processing.
However, these state-of-the-art frameworks exhibit notable limitations. They fail to fully parallelize various computing operations, resulting in instances of sequential execution. 
The bottleneck of computing throughput lies in data transmission between CPU and GPU, impacting both environment interaction and network update throughput.
In multi-GPU scenarios, limited optimizations cater to the characteristics of RL algorithms.
Furthermore, manual setting of parallelization hyperparameters in many frameworks hinders convenient deployment on diverse computing devices.

Our contribution addresses these bottlenecks by proposing methods to enhance throughput.
This includes implementing a shared memory system to reduce data transfer during environment interaction and utilizing ``Actor-Critic'' Model Parallelism to mitigate data transfer issues during network updates. Our hyperparameter adaption model ensures optimal configuration across different computing devices, collectively improving system efficiency.
Our objective is to construct a high-throughput RL framework fully parallelizing experience sampling, network update, and performance evaluation functions (Fig.~\ref{fig_spreeze}).
Focusing on efficient RL training on a desktop, with large-batch training, maximizes hardware capabilities (GPU, CPU, memory, and hard disk). 
Operations like shared memory on a single desktop expedite data transmission, and an efficient parallel design caters to most RL training tasks.
In contrast to frameworks sacrificing calculation efficiency for algorithm adaptability, ours prioritizes efficiency, specifically tailored for mainstream off-policy single-agent RL algorithms, resulting in a significant 73\% reduction in average training time.

The main contributions of our work are as follows:

\begin{itemize}

    \item We have developed a multiprocess parallel RL framework that maximizes the performance of GPU, CPU, memory, hard disk and other hardware components. High throughput is achieved through multiple parallel sampling processes and the unified large-batch network update. 
    
    \item Our research reveals the impact and underlying principle of various parallelism hyperparameter settings on training speed, and achieves the balance between experience sampling efficiency and network update efficiency based on the performance of hardware equipment.
    
    \item In consideration of the ``Actor-Critic'' RL algorithm feature, our framework designs a method to update the Actor network and Critic network concurrently with dual GPUs in order to further enhance the throughput.
    
\end{itemize}

We introduce the work related to high-speed RL training in the following Section~\ref{related_works}.
In Section~\ref{framework}, our RL framework and methods are introduced in detail.
Section~\ref{experiment} conducts experiments to verify the performance of our framework and reveal some principles of high-throughput parallel RL calculations.
Finally, we summarize the whole paper in Section~\ref{conclusion}.

\section{Related works}

\label{related_works}

The RL community has been dedicated to improving the speed of RL training and has proposed a variety of RL acceleration methods. 
Although methods such as those the Zero series researches \cite{silver2017mastering, schrittwieser2020mastering} utilize a large number of GPUs for computation, where GPUs are mainly used for parallel Monte Carlo tree search and evaluation of the large value-network, the methods are not universally applicable to standard RL algorithms.

\subsection{Parallel Reinforcement Learning}

At present, the training process of most RL research is completed in simulation, and then the trained network can be transferred to the real world for application.
In simulation, experience sampling is relatively fast and cheap, unlike realworld training that pays much attention to sample efficiency. Therefore, multiple sampling programs can be established for parallel sampling to obtain a better exploration effect.

A3C \cite{mnih2016asynchronous} is a distributed and asynchronous training method widely used at early stages. In its asynchronous distributed process, it not only performs experience sampling operations, but also performs network update operations. However, the experience provided by a single sampler is insufficient, and it is difficult to perform effective large-batch training to make full use of the GPU, which leads to low network updates and experience sampling throughput.  
Ape-X \cite{horgan2018distributed} stores the experience collected by multiple samplers in a unified experience pool for learners to update the network efficiently in large-batch. 
In IMPALA \cite{espeholt2018impala}, the actors are also only responsible for experience sampling, but directly transmit experience to the learner to update the network without using the experience buffer. 
In addition, DD-PPO \cite{wijmans2019dd} is a parallel algorithm focusing on multi-GPU optimization. 
The seed rl method \cite{espeholt2019seed} proposed by Google is based on the traditional V-trace and Q-learning algorithms, and is not suitable for the current mainstream executor-reviewer algorithms such as SAC \cite{haarnoja2018soft} and DDPG \cite{lillicrap2015continuous}.

RLlib \cite{liang2018rllib} is a well-known RL framework based on Ray \cite{moritz2018ray} to achieve parallelization, which implements many RL algorithms including multi-agent RL and model-based RL. 
Acme \cite{hoffman2020acme} is a simplified novel RL framework proposed by DeepMind, which focuses on parallelization for high-speed training and supports multiple RL simulation environments. 
In addition, rlpyt \cite{stooke2019rlpyt} is a small and medium-sized deep RL framework based on PyTorch. Although it performs parallelization operations and also claims high throughput, quantitative throughput experiments are lacking.

Overall, although the existing RL frameworks have designed a variety of parallel experience sampling and learning operations, they do not pay special attention to their data throughput, which is actually not very high.
Compared with the existing work, our framework is more thoroughly parallelized. Not only is the experience sampling parallelized, but also the network update, performance testing, and visualization functions are separated into dedicated processes to make full use of computer hardware equipment. 

\begin{figure}[t]
	\centering
	\subfigure[Data parallelism]{
		\includegraphics[width=0.99\linewidth]{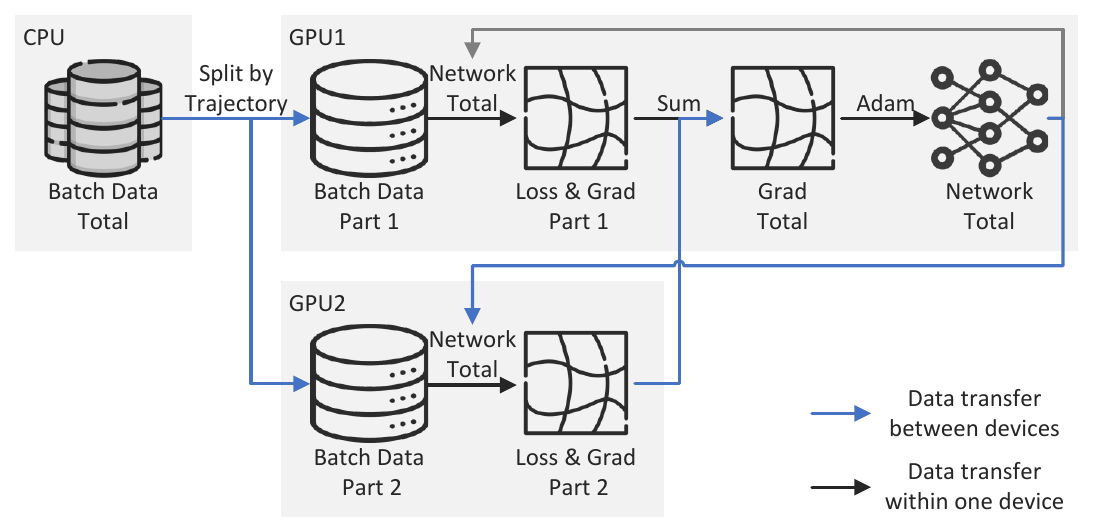}  
	}
	\subfigure[``Actor-Critic'' Model parallelism]{
		\includegraphics[width=0.99\linewidth]{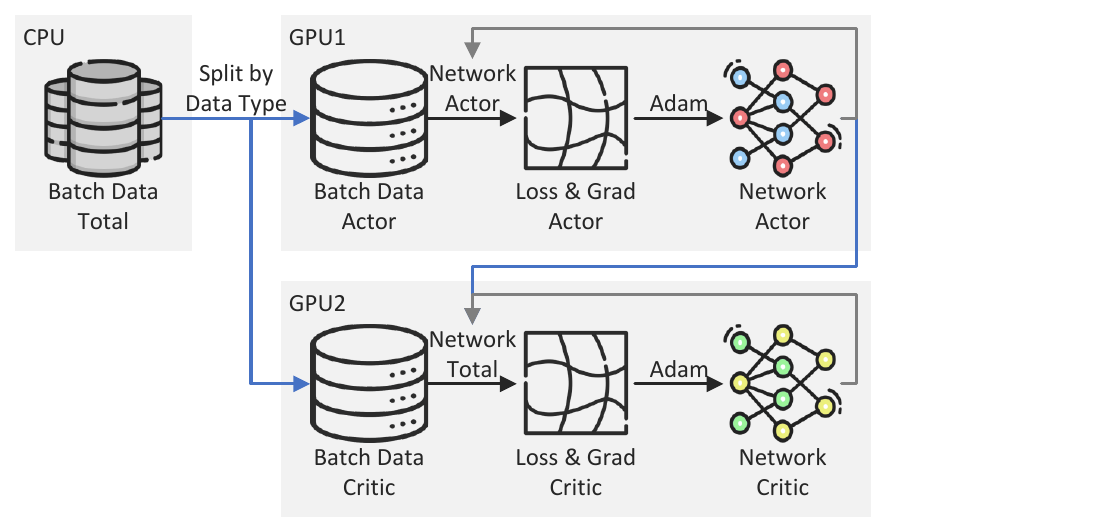}
	}
	\caption[]{\textbf{Comparison of multi-GPU parallelization methods for RL training frameworks.}}
	\label{fig_multi_gpu}
\end{figure}

\subsection{Large-Batch Training}  
Large-batch training is also an effective way to improve training effect and speed.
In the accelerated methods of deep reinforcement learning (DRL) \cite{stooke2018accelerated}, it has been found that training with a batch size much larger than the standard can obtain relatively good training results.
Similarly, in the shallow updates study of DRL \cite{levine2017shallow}, the use of a large batch size of up to 4096 in the last layer of the network can significantly improve the performance.
\cite{hoffer2017train} study the use of large batch size for machine learning and find that it can effectively improve the generalization ability of the network, thereby improving the training effect.
In addition, some adaptive batch size methods for safety policies \cite{papini2017adaptive} or for continuous actions \cite{han2017amber} have also been proposed.
In our framework, setting a large batch size for large-batch training can effectively improve the training performance. 

\subsection{Data Parallelism and Model Parallelism}
Data parallelism and model parallelism are two commonly used techniques in distributed machine learning systems. 
Data parallelism involves distributing the training data across multiple devices and training identical copies of the model on each device \cite{dean2012large, ying2018image}. 
Model parallelism involves dividing the model into smaller sub-models and running each sub-model on a different device. Research works have explored various optimization techniques to improve the efficiency and scalability of these techniques, including gradient compression, network topology optimization, and dynamic load balancing \cite{huang2019gpipe}. 
For RL with a small amount of observed feature data, data parallelism will frequently transfer network gradients and parameters between different hardware devices, resulting in additional communication overhead and inefficient training. Consequently, we design an expandable model parallel method based on the characteristics of the actor-critic network structure of RL.

\section{High-throughput Framework Design}


\label{framework}

Our proposed Spreeze framework is a multi-process RL framework, which asynchronously parallelizes various time-consuming aspects of RL training, thereby maximizing the utilization of GPU, CPU, memory, hard disk and other hardware devices. 
As shown in Fig.~\ref{fig_spreeze}, our framework includes multiple environment interaction processes, a network update process, a test process, and a visualization process.

The asynchronous environment interaction Process orchestrates concurrent interactions between agents and the environment, generating diverse experience data. This data is then transmitted to the network update module, where large-batch training and actor-critic model parallelism optimize learning efficiency. The variable transmission by RAM and SSD ensures timely and accurate data sharing between processes, utilizing shared memory for experience data and solid-state drive storage for network weights. Hyperparameter adaptation dynamically adjusts key training parameters based on hardware performance, optimizing resource utilization in a feedback loop with the asynchronous interaction and network update processes. Validation and visualization processes provide insights into the algorithm's performance, contributing to a comprehensive understanding of the trained model.

\subsection{Asynchronous Environment Interaction Process}
The asynchronous environment interaction method is a key component of our high-throughput parallel RL framework. This method enables multiple agents to learn and interact with an environment simultaneously, allowing for more efficient and effective learning. This approach improves both sample efficiency and learning speed by leveraging parallel computation resources.
Specifically, the actors include experience sampling processes, test processes, and visualization processes.

\subsubsection{Experience Sample}

Our framework incorporates asynchronous experience sampling as a core component of training. The process involves generating rich experience data by allowing multiple agents to interact with the environment simultaneously. The parallelization of experience sampling is achieved using multi-process technology, where a large number of empirical sampling processes are generated concurrently. The maximum number of processes generated is dependent on the number of cores of a given CPU.


\begin{figure}[t]
	\centering
    \includegraphics[width=0.99\linewidth]{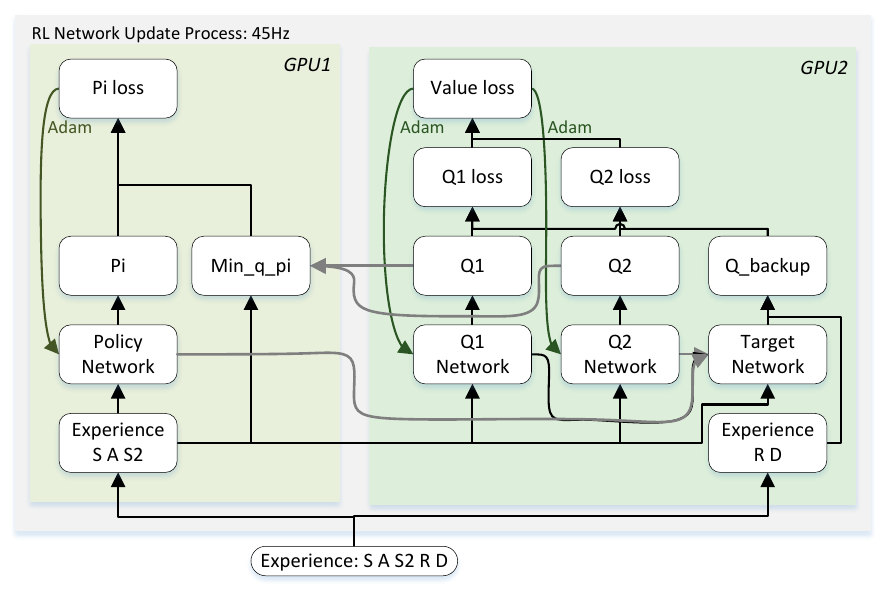}
	\caption[]{\textbf{Network update architecture.} Two GPUs respectively update the actor policy network and the critic value network with as little data transmission as possible. The shown throughput takes the PyBullet Walker-2d task as an example.}
	\label{fig_update}
\end{figure}

During training, N sampling processes are continuously running and interacting with the environment to generate experience trajectories. The policy network generates actions by forward propagation based on the collected data from the multiple sampling processes. This results in a more efficient and effective learning strategy, as the network updates are based on rich and diverse data.


Each experience sampling process can fully utilize the computing resources of one CPU core, allowing for maximum use of available resources. However, since the CPU also performs other tasks, such as test verification, visualization, or data transfer, not all experience sampling processes can run continuously. To ensure optimal performance, the running number N of experience sampling processes is dynamically adjusted based on the CPU load during training.


The adjustment strategy is implemented to ensure that there is no wastage of resources and that the available computing power is effectively allocated to each process. This results in a more efficient and faster training process, allowing for the generation of high-quality RL models. For the specific adjustment strategy, see Section 3.4. In conclusion, the asynchronous experience sample actor method is a crucial component of our high-throughput framework. It allows for the generation of rich experience data by enabling multiple agents to interact with the environment simultaneously. The process is optimized through network updates based on collected data from the multiple sampling processes.

\subsubsection{Validation and Visualization}

In addition to the asynchronous experience sampling process, our high-throughput parallel RL framework also incorporates the asynchronous validation and visualization sample actor method to further enhance the training process. The test process generates an episode return curve to obtain a dense and accurate reward curve, while the visualization process displays the process of interaction between the algorithm and the environment to showcase the learned strategy.


These processes are variants of the experience sampling process, but they do not randomize the actions output by the policy network and do not transmit experience to the update process. As the frame rate of the visualization process is much lower than that of the test process, the two are not integrated into one process. This approach results in a more efficient and effective training process, leading to the generation of high-quality RL models.

\subsection{Network Update}
\subsubsection{Large-Batch Training}

Large-batch training is a method of RL that utilizes parallelization to accelerate the algorithm. Due to hardware limitations, it is challenging to increase the frequency, so large-batch parallelization is the core of algorithm acceleration.
In our framework, the experience collected from multiple experience sampling processes is transmitted to a single network update process for parallelization by GPU. To ensure robustness in RL training, a large batch size is set for parallelization. A large batch size can achieve high training efficiency and keep the training curve stable. However, the batch size is limited by the GPU's memory capacity and computing power.

To address this limitation, our framework adaptively selects the largest possible batch size based on the GPU's performance to achieve high-throughput network updates. The hyperparameter determination strategy is described in Section 3.4. By selecting the optimal batch size, we can balance the computational efficiency and stability of the training process, resulting in a faster and more stable RL training process.

\subsubsection{Actor-Critic Model Parallelism}

In addition to large-batch training, the parallelization of multiple GPUs for network updates is another method used in our framework for RL training. Since most RL algorithms use the ``Actor-Critic'' dual-network architecture, which consists of an actor network responsible for selecting actions and a critic network responsible for evaluating the quality of actions, an architecture is designed in which actor and critic networks are distributed on two GPUs for independent updates, as shown in Fig. \ref{fig_multi_gpu}(b). 
Our approach is different from the data parallel method used in supervised learning (Fig. \ref{fig_multi_gpu}(a)) because the network model of RL tasks is usually less complex, and it is not cost-effective to distribute gradient calculations to multiple GPUs for calculation.


To specifically illustrate how the framework parallelizes the multi-GPU network computing graph model, we take the SAC \cite{haarnoja2018soft}  algorithm as an example. The specific calculation diagram is shown in Fig.~\ref{fig_update}. In this method, each GPU is responsible for a part of the calculation and minimizes the amount of data communication between GPUs. GPU0 is mainly responsible for updating the policy network, while GPU1 is mainly responsible for updating the value network. Experience data, including state $s$, action $a$, next state $s2$, reward $r$, and done flag $d$, is distributed to two GPUs as required by the network model. The reward and done experiences are allocated to GPU1 because they are only used to calculate the value network loss, while other experiences are allocated to GPU0.


Since the network architecture draws on the double-Q algorithm to prevent overestimation, there are two value networks Q1 and Q2. They are updated by GPU1 together with the target network, while GPU0 is responsible for calculating the loss of the policy network and performing gradient descent updates. The parallelization of the multi-GPU model reduces the time required for network updates and accelerates the training process.
The Actor-Critic Model Parallelism method for RL in our framework distributes the actor and critic networks on two GPUs for independent updates, minimizing the amount of data communication between GPUs. This approach reduces the time required for network updates and accelerates the training process, making it a valuable addition to the large-batch training method for parallelization.

\begin{figure}[t]
	\centering
	\subfigure[Partial asynchronous parallelization without shared memory]{
		\includegraphics[width=0.99\linewidth]{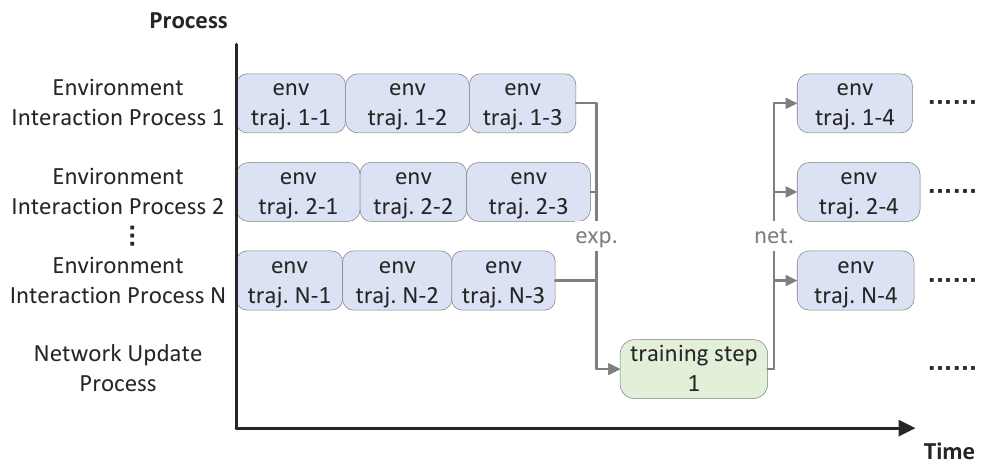}  
	}
	\subfigure[Full asynchronous parallelization with shared memory]{
		\includegraphics[width=0.99\linewidth]{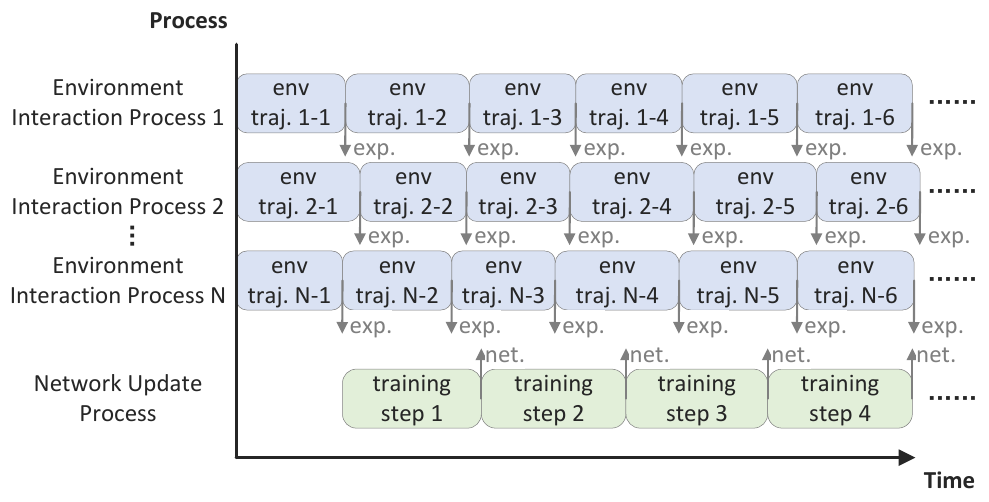}
	}
	\caption[]{\textbf{Multi-process parallel comparison of whether to use shared memory.} ``exp.'' represents the experience data generated from the environment interaction process; and ``net.'' represents the network parameters generated by the network update process.}
	\label{fig_process}
\end{figure}

\subsection{Variable Transmission}
\subsubsection{Transmission Framework}

Our high-throughput parallel RL training framework relies on efficient variable transmission between processes to achieve optimal performance. This framework involves two main types of data transmission: sampled experience and neural network weights. Since the experience sampling rate is much slower than the network update rate, multiple experience sampling processes are established to provide data for a single network update process. However, this leads to relatively large data reception pressure on the network update process, which necessitates high-speed transmission for the experience data.


On the other hand, the network weight is a Tensor variable \cite{abadi2016tensorflow} in our framework that poses a challenge for transmission using traditional Queue or shared memory operations. Nevertheless, due to its infrequent transmission requirements and the need to save checkpoint points periodically, the network weight can be transmitted using solid-state drive (SSD) storage. By saving and reading the variable in each process, we can ensure that the network weight is transmitted accurately and efficiently, without causing any bottlenecks in the framework. 
Our variable transmission framework helps to optimize the use of computing resources and improve the training speed and accuracy of our RL models.

\subsubsection{Replay Buffer Shared Memory}

To achieve efficient experience transfer and high-speed training in parallel RL, conventional methods such as Queue or Pipe operations used for data transfer between processes can be problematic. They consume a lot of time for data dump, causing delays in the training process, and a large queue size can cause data lag, affecting the training performance. To overcome these issues, we use shared memory technique \cite{ahn2018soft} to transfer experience data.

The shared memory method enables direct updating of the experience pool for network updates without consuming the time of the network update process. This approach allows experience obtained from sampling to be used for network updates as soon as possible, without waiting for the queue to be fully collected. Locking mechanisms are used to prevent data confusion when accessing the experience in shared memory. 
If the queue method is used to transmit data, it needs to occupy the time of both the sending process and the receiving process, which makes the existing RL framework adopt a partial parallelization mode as shown in Figure 4a. This requires the environment interaction process and the network update process to centrally agree on a time for data transmission, resulting in a large amount of wasted waiting time. 

The shared memory method we adopt does not take up the time of receiving memory, so that each process can perform calculations uninterrupted and send data to other processes in a timely manner, thus greatly improving hardware usage and training efficiency.
Shared memory technology facilitates the transmission of data without causing interruptions to the processing time of the receiving process. This enhancement significantly improves the efficiency of a fully asynchronous experience sampling, as illustrated in Fig. \ref{fig_process}.

Compared to the Queue method, the shared memory technique offers a significant improvement in experience transfer frequency, achieving 10 Hz without wasting update process time. In contrast, the Queue method only achieves a transfer frequency of 0.2 Hz, wasting about 20\% of the update process time. Our experiments in section~5 demonstrate the efficacy of this approach.

\begin{table*}[]
\centering
\caption{\textbf{Comparison of time to solve between different frameworks}}
\begin{tabular}{crclrclrclrclc}
\hline
Env\textbackslash{}Framework & \multicolumn{3}{c}{Spreeze(Ours)} & \multicolumn{3}{c}{RLlib} & \multicolumn{3}{c}{ACME} & \multicolumn{3}{c}{rlpyt} &  \begin{tabular}[c]{@{}c@{}}Time \\ Save\end{tabular} \\ \hline
Pendulum                             & \textbf{16.3}        & \textbf{±}      & \textbf{1.4}        & 209.4     & ±   & 1.1     & 173.6    & ±   & 22.5    & 158.3    & ±   & 20.9     & 89.7\%      \\
HalfCheetah                          & \textbf{215.3}       & \textbf{±}      & \textbf{49.8}       & 4337.2    & ±   & 1124.5  & 700.2    & ±   & 126.2   & 1754.2   & ±   & 96.9     & 69.3\%      \\
Walker                               & \textbf{220.9}       & \textbf{±}      & \textbf{5.5}        & 1884.6    & ±   & 1798.1  & 3485.1   & ±   & 677.1   & 748.4    & ±   & 78.2     & 70.5\%      \\
Ant                                  & \textbf{414.6}       & \textbf{±}      & \textbf{104.8}      & 997.8     & ±   & 630.3   & 614.2    & ±   & 100.8   & 582.4    & ±   & 78.8     & 28.8\%      \\
Humanoid                             & \textbf{668.2}       & \textbf{±}      & \textbf{79.3}       & 3861.2    & ±   & 819.2   & \multicolumn{3}{c}{-}    & 8215.4   & ±   & 1930.0   & 82.7\%      \\
HumanoidRunFlag                      & \textbf{904.1}       & \textbf{±}      & \textbf{66.4}       & 19461.3   & ±   & 2019.0  & \multicolumn{3}{c}{-}    & \multicolumn{3}{c}{-}     & 95.4\%      \\ \hline
average                              & \textbf{487.9}       & \textbf{±}      & \textbf{69.6}  & 6150.3       & ±      & 1132.3 & \multicolumn{3}{c}{-}   & \multicolumn{3}{c}{-}                                                                                             & 72.7\%      \\ \hline
\end{tabular}
\begin{tablenotes}
	\footnotesize
	\item * The unit of time results in the table is "seconds". Bold numbers indicate time results for the least time-consuming frame in each environment. "Time Save" refers to the percentage of time our framework saves compared to the best among the three other frameworks. The average ``Time Save'' is calculated as the mean value across five environments.

\end{tablenotes}
\label{tab:solve}
\end{table*}

\begin{figure*}[t]
	\centering
	\subfigure[Pendulum-v0]{
		\includegraphics[width=0.23\linewidth]{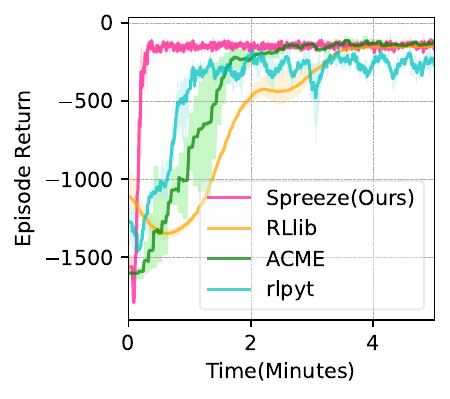}  
	}
	\subfigure[Walker2D]{
		\includegraphics[width=0.23\linewidth]{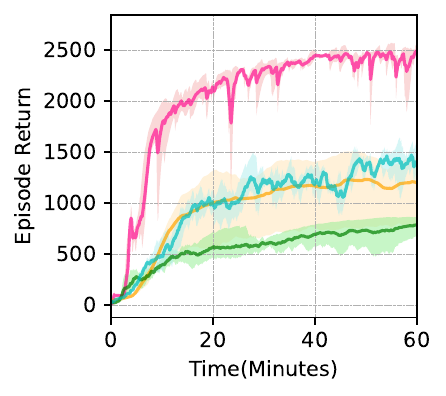}
	}
	\subfigure[Ant]{
		\includegraphics[width=0.23\linewidth]{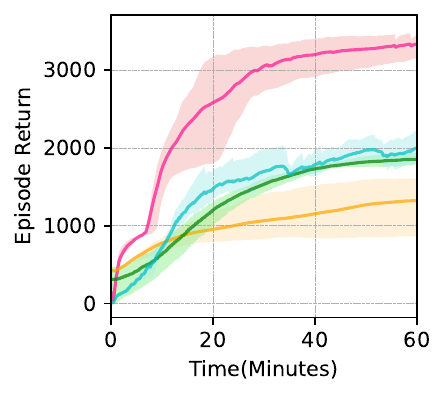}
	}
	\subfigure[Humanoid]{
		\includegraphics[width=0.23\linewidth]{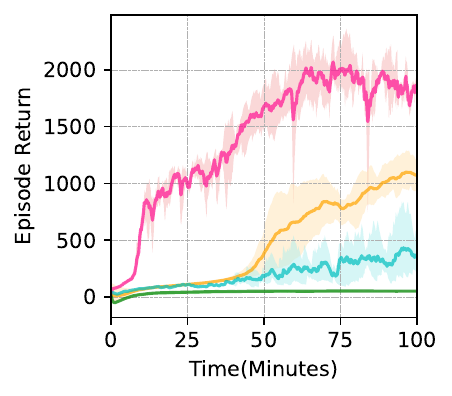}
	}
	\caption[]{\textbf{Performance comparison of different frameworks in different environments.} We have performed a detailed hyperparameter search for each framework to ensure that the best performance of each framework can be basically achieved. Each curve represents the average of five random seed experiments.}
	\label{fig_result}
\end{figure*}

\subsection{Hyperparameter Adaptation}
\subsubsection{Hyperparameter Impact}

Hyperparameters play a crucial role in high-throughput hyperparameter adaptation parallel RL training as they significantly impact the degree of parallelization. Setting hyperparameters manually can be challenging and time-consuming, and may not achieve optimal training results. Therefore, in our parallel framework, we have designed a parameter adaptation function that adjusts the hyperparameters based on hardware performance to achieve optimal parallelization results automatically.


Two key hyperparameters that affect parallelization performance are batch size and the number of sampling processes. 
Batch size mainly affects the GPU, as it determines the number of experience frames included in the network update. When the GPU occupancy rate is low, increasing the batch size does not affect the network update frequency but increases the number of experience frames included in the network update. On the other hand, when the GPU occupancy rate is close to saturation, increasing the batch size becomes difficult to further increase the network update frame rate, but it makes the network update frequency decrease. The number of sampling processes, on the other hand, impacts the CPU performance. Increasing the number of sampling processes can enhance the rate of experience sampling, but if too many processes occupy the CPU, it can reduce the efficiency of the network update process. 

In general, the high -parameter adaptation function based on hardware performance requires our parallel framework to optimize the key super parameters, so as to obtain better parallelism and performance to achieve parallel RL training adapted to high-throughput supers tipping.

\subsubsection{Adaptation Strategy}

Current RL hyperparameter adaptation methods employ population evolution algorithms to periodically adjust hyperparameters, such as the learning rate, at fixed intervals during network training. In our framework, we focus on adapting two critical hyperparameters: the batch size and the number of experience sampling processes. We optimize the number of experience sampling processes to maximize the throughput of experience sampling, and we adapt the batch size to optimize the data throughput for network updates. Notably, these two hyperparameters are largely independent, allowing us to optimize them individually.

The number of experience sampling processes must be an integer, with its optimal value often aligning closely with the available CPU cores, typically ranging from a few to several dozen. When it comes to batch size, the choices are generally restricted to geometrically increasing values. Consequently, we can achieve adaptive functionality through a straightforward search of hyperparameters using enumeration methods. However, we have employed additional insights to expedite the search process. Specifically, when the number of experience sampling processes is too low, it results in underutilization of the CPU resources. Similarly, if the batch size is set too small, it leads to underutilization of the GPU resources. Furthermore, it is worth noting that the relationship between experience sampling throughput and the number of experience sampling processes exhibits a convex behavior, as does the relationship between network update data throughput and batch size.

In summary, our hyperparameter adaptation strategy involves monitoring the CPU usage to ascertain whether it is excessive or insufficient, and then adjusting the number of experience sampling processes accordingly while meeting the memory capacity constraints until the experience sampling throughput reaches an optimal point. We also monitor GPU usage, making corresponding adjustments to the batch size to maximize network update throughput without exceeding the GPU capacity.

\section{Performance Evaluation}

\label{experiment}

In this section, numerical studies are conducted to evaluate the performance of our high-throughput RL framework. We compare the throughput, training effect, and hardware utilization of our framework to those of other prevalent RL parallel frameworks, and reveal the fundamental principles for achieving high-throughput parallel training.

\subsection{Experiment Setup}
Our framework is compatible with any standard OpenAI Gym \cite{brockman2016openai} environment, which is a mainstream RL training environment.
In our experiments, we choose the PyBullet \cite{coumans2016pybullet} robot control simulation platform, which is widely used as a RL robot control benchmark and is based on Gym.
We chose three robot models with different difficulty levels, from easy to difficult, Walker2D, Ant, and Humanoid.
In addition, in order to test the training efficiency of each RL framework in a relatively simple environment, we also select the Pendulum-v0 environment based on OpenAI gym \cite{brockman2016openai} for experiments.
The main hardware platform used in the experiment includes a 12-core AMD 5900X CPU, NVIDIA 1060 GPU, and 32G RAM, which are common personal desktop configurations.
Our framework will automatically adjust parameters such as batch size and the number of processes based on hardware performance.

\begin{table*}[htbp]
\centering
\caption{\textbf{Comparison of hardware usage and throughput between different frameworks}}
\begin{tabular}{llllll}
\hline
Framework\textbackslash{}Index & \begin{tabular}[c]{@{}l@{}}CPU\\ Usage $\uparrow$\end{tabular} & \begin{tabular}[c]{@{}l@{}}Sampling \\ Frame\\ Rate (Hz) $\uparrow$\end{tabular} & \begin{tabular}[c]{@{}l@{}}GPU\\ Usage $\uparrow$\end{tabular} & \begin{tabular}[c]{@{}l@{}}Network Update\\ Frame Rate (Hz) $\uparrow$\end{tabular} & \begin{tabular}[c]{@{}l@{}}Network Update\\ Frequency (Hz) $\uparrow$\end{tabular} \\ \hline
Spreeze(Ours)                           & \textbf{75\%}                                                & \textbf{15342}                                                    & \textbf{82\%}                                       & \textbf{3.7E+5}                                                     & 45.2                                                               \\
Spreeze-BS128(Ours)                     & \textbf{75\%}                                                & \textbf{15564}                                                             & 60\%                                                & 4.2E+4                                                              & \textbf{330.3}                                                     \\
\hline
RLlib-APEX-BS128               & 64\%                                       & 4132                                                             & 32\%                                                & 3.3E+4                                                              & 257.6                                                              \\
RLlib-APEX-BS4096              & 63\%                                                & 4513                                                             & 30\%                                                & 3.6E+4                                                              & 8.8                                                                \\
RLlib-PPO-CPU-BS128            & 25\%$\sim$100\%                                     & 2244                                                             & 0\%                                                 & 2244                                                                & 17.5                                                               \\
RLlib-PPO-CPU-BS8192           & 25\%$\sim$100\%                                     & 2204                                                             & 0\%                                                 & 2204                                                                & 0.3                                                                \\
RLlib-PPO-GPU-BS128            & 15\%$\sim$100\%                                     & 1268                                                             & 42\%                                                & 1268                                                                & 9.9                                                                \\
\hline
ACME-BS256                     & 11.7\%                                              & 420                                                              & 18.7\%                                              & 4.9E+3                                                              & 0.593                                                              \\
ACME-BS8192                    & 10.1\%                                              & 590                                                              & 11.2\%                                              & 1.1E+4                                                              & 39.8                                                               \\ 
\hline
rlpyt-BS128                    & 52\%                                              & 11080                                                              & 45\%                                              & 1.3E+4                                                              & 103.6                                                               \\ 
rlpyt-BS512                    & 60\%                                              & 14898                                                              & 35\%                                              & 4.5E+4                                                              & 88.7                                                               \\ 
\hline
\end{tabular}
\begin{tablenotes}
	\footnotesize
	\item * ``$\uparrow$'' indicates that the higher the performance index, the better. The bold values represent the best performance in each index. ``BS'' represents the batch size.
\end{tablenotes}
\label{tab_framework}
\end{table*}

\subsection{Results}

\subsubsection{Parallel Framework Performance Comparison}
First, we conduct performance comparison experiments between different frameworks.
We choose the widely used RLlib \cite{liang2018rllib}, Acme \cite{hoffman2020acme} and rlpyt \cite{stooke2019rlpyt} as the comparison frameworks. 
RLlib implements distributed training based on the Ray parallel AI framework \cite{moritz2018ray}.
The rlpyt framework uses the PyTorch library to unify the three types of algorithms: deep Q-learning, policy gradients, and Q-value policy gradients.
Acme, on the other hand, focuses on enabling scalable distributed RL training at different execution scales.

\begin{figure*}[htbp]
	\centering
	\subfigure[Experience Transition Method]{
		\includegraphics[width=0.31\linewidth]{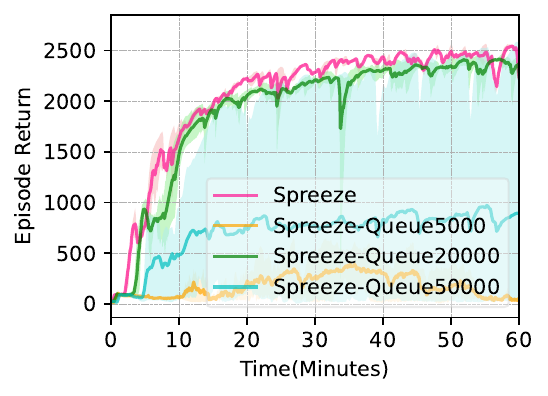}  
	}
	\subfigure[CPU Hardware Device]{
		\includegraphics[width=0.31\linewidth]{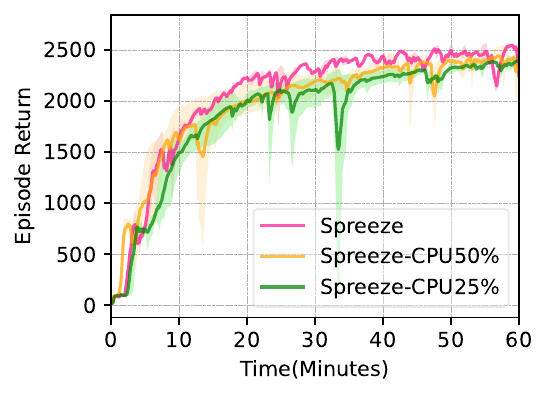}
	}
	\subfigure[GPU Hardware Device]{
		\includegraphics[width=0.31\linewidth]{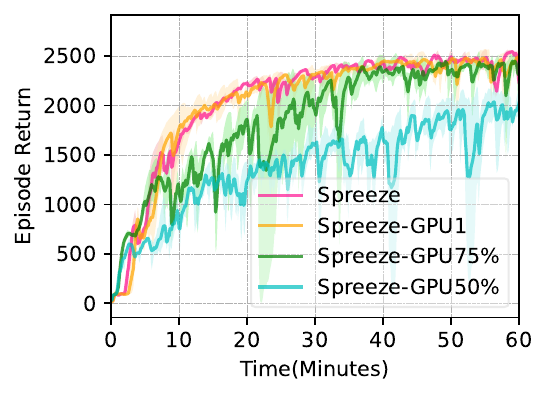}
	}
	\caption[]{\textbf{Ablation Experiment.} (a) Performance comparison between standard shared memory experience transfer and queue experience transfer with different queue size. (b) Ablation experiment limited to use only 50\% and 25\% CPU hardware resources. (c) Ablation experiment limited to use only a single or 75\% or 50\% GPU hardware resources. Five random seed trainings are performed under each experimental condition.}
	\label{fig_aba}
\end{figure*}

To facilitate a comprehensive comparison of parallelization efficiency, meticulous parallelization settings were configured for each framework.
Given the diverse approaches to parallelization in each framework, distinct optimal RL algorithms and parameters were identified for the execution of experimental tasks.
Initially, our intention was to compare identical mainstream RL algorithms, such as SAC, across different frameworks.
However, it was observed that RLlib, despite parallelizing the SAC algorithm, lacked the implementation of a unified experience playback mechanism similar to ours, resulting in suboptimal overall training efficiency.
In the case of RLlib, the parallel APE-X algorithm exhibited superior performance compared to the SAC algorithm. Consequently, we opted for the APE-DDPG variant algorithm from RLlib for our continuous motion control experiments.
Subsequently, the upcoming experiments will encompass hyperparameter searches for various frameworks to attain optimal performance for each and enable a fair and informed comparison.
As delineated in Table \ref{tab_framework}, the hyperparameter search for Batch size spanned settings from 128 to 8192 across different frameworks, aiming to identify the maximum data throughput and achieve the highest training efficiency.
In specific scenarios, such as when training the PyBullet task in our framework, the optimal batch size was determined to be approximately 8192, while the Acme framework exhibited optimal performance with a batch size of 256.
For PyBullet tasks, we selected the most suitable algorithms from RLlib, Acme, and rlpyt—PPO \cite{schulman2017proximal}, D4PG \cite{barth2018distributed}, and SAC \cite{haarnoja2018soft}, respectively.
Given the cost-effectiveness and rapidity of experience sampling in simulation, the focus of the parallelization framework is predominantly on time-efficiency rather than sample-efficiency \cite{hoffman2020acme}.
Furthermore, due to varying levels of support for multi-GPU training among frameworks, we opted to employ a single GPU when conducting comparative analyses across different frameworks.

As shown in Fig.~\ref{fig_result}, the horizontal axis is the training time and the vertical axis is the test episode return.
Following the setting of \cite{liang2018rllib}, we sort out the time required for each framework to solve different tasks as shown in Tab. \ref{tab:solve}.
The target returns for Pendulum, HalfCheetah, Walker, Ant, Humanoid, and HumanoidRunFlag are -200, 800, 850, 850, 1800, and 100, respectively.
The results show that our framework is significantly better than mainstream RL parallel frameworks such as RLlib, Acme and rlpyt in terms of training efficiency on tasks of various difficulties. 
To achieve the same control performance, our training time is reduced by about 73\% compared to other frameworks.

\subsubsection{Hardware Usage and Throughput Analysis}
The core of improving the training efficiency of the RL framework is to make full use of computing hardware resources and increase data throughput, so we conduct experiments and analysis on the hardware usage and throughput.
We take the Walker2D environment as an example to analyze the impact of different parameter settings in our framework on the above indicators, and the results are shown in Table~\ref{tab_para}. 
The index of hardware throughput includes CPU usage, sampling frame rate, GPU usage, network update frame rate, network update frequency, experience transfer cycle, experience transmission frame rate, and experience transmission loss. 
Among them, the network update frame rate refers to the multiplication of network update frequency and batch size. 
In order to facilitate the analysis of GPU occupancy, the experiments in this section still only use a single GPU.
 
Next, we will also conduct a comparative analysis of hardware usage and throughput among different frameworks. 
Various algorithms and parameters under the RLlib, Acme, rlpyt and Spreeze frameworks have been tested, fully demonstrating the performance of each framework.
The results are shown in Table~\ref{tab_framework}.
It can be found that the throughput of our framework is significantly larger than that of other frameworks in both experience sampling and network update.
In particular, the network update frame rate of our framework is an order of magnitude higher than that of other frameworks, which is considered the key to our framework to train RL agents faster.
In other frameworks, the problem of low network update frame rate and low GPU utilization is common. After trying to increase the batch size for them, the network update frame rate can only be slightly increased, but this significantly reduces the network update frequency and worsens the training effect. Increasing the batch size in the rlpyt framework will greatly occupy memory, and 32G memory only supports increasing the batch size to 512.
In addition, the CPU and GPU occupancy rates of our framework are also the highest among all the frameworks. However, our occupancy rates have not reached 100\% because as shown in section 3.3 and Table 2, fully occupying the CPU or GPU will bring a series of adverse effects.

\begin{table*}[tb]
\centering
\caption{\textbf{The impact of hyperparameters in our framework on hardware usage and throughput}}
\begin{tabular}{llllllll}
\hline
Framework\textbackslash{}Index & \begin{tabular}[c]{@{}l@{}}CPU\\ Usage $\uparrow$ \end{tabular} & \begin{tabular}[c]{@{}l@{}}Sampling \\ Frame\\ Rate (Hz) $\uparrow$\end{tabular} & \begin{tabular}[c]{@{}l@{}}GPU\\ Usage $\uparrow$\end{tabular} & \begin{tabular}[c]{@{}l@{}}Network \\ Update Frame\\ Rate (Hz) $\uparrow$\end{tabular} & \begin{tabular}[c]{@{}l@{}}Network\\ Update\\ Frequency\\ (Hz) $\uparrow$\end{tabular} & \begin{tabular}[c]{@{}l@{}}Experience\\ Transfer\\ Cycle (s) $\downarrow$\end{tabular} & \begin{tabular}[c]{@{}l@{}}Experience\\ Transmission\\ Loss $\downarrow$\end{tabular} \\ \hline
Spreeze                           & 75\%                                                & 15342                                                             & 82\%                                                & 3.7E+5                                                              & 45.2                                                               & 0                                                                          & 3\%                                                                     \\
Spreeze-BS32768                   & 75\%                                                & 16207                                                             & 86\%                                                & \textbf{4.1E+5}                                    & \underline{12.4}                                                                       & 0                                                                          & 3\%                                                                     \\
Spreeze-BS128                     & 75\%                                                & 15564                                                             & \underline{60\%}                                       & \underline{4.2E+4}                                                     & \textbf{330.3}                                & 0                                                                 & 3\%                                                                                                    \\
Spreeze-SP16                      & \textbf{90\%}                                        & \textbf{17122}                                                   & 79\%                                                & \underline{3.2E+5}                                                     & \underline{39.1}                                                     & 0                                                                          & \underline{10\%}                                                            \\
Spreeze-SP2                       & \underline{21\%}                                       & \underline{3300}                                                  & 84\%                                                & \textbf{3.9E+5}                                                   & \textbf{48.1}                                                        & 0                                                                          & \textbf{2\%}                                                               \\
\hline 
Spreeze-QS5000                    & \underline{50\%}                                                & \underline{8044}                                                             & 81\%                                                & \underline{2.2E+5}                                                     & \underline{26.9}                                                     & \underline{5.0}                                                                & \underline{45\%}                                                          \\
Spreeze-QS20000                           & \underline{50\%}                                                & \underline{8672}                                                             & 83\%                                                & \underline{3.2E+5}                                                              & \underline{39.1}                                                               & \underline{18.1}                                                                          & \underline{16\%}                                                                     \\
Spreeze-QS50000                   & \underline{49\%}                                                & \underline{8630}                                                             & 82\%                                                & 3.6E+5                                                      & 44.5                                                   & \underline{46.0}                                                                 & \underline{9\%}                                                               \\
\hline
\end{tabular}
\begin{tablenotes}
	\footnotesize
	\item * ``$\uparrow$'' indicates that the higher the performance index, the better, while ``$\downarrow$'' indicates that the lower the index, the better. The boldface indicates performance improvement compared to the standard configuration, and the underline indicates performance deterioration. ``BS'' represents the batch size, ``SP'' represents the number of sample processes, and ``QS'' represents the queue size.
\end{tablenotes}
\label{tab_para}
\end{table*}

\subsubsection{Ablation Experiments}

Next, we conduct experiments from two aspects of parallelization technique ablation and hardware device limitations.
In the experiments, we use the dual GPU independent update ``Actor-Critic'' network technique described in section 3.3, and our experiment performs five random seed trainings for each setting in the Pybullet 2D humanoid robot control environment.

For parallelization technology, we test the final performance difference between using shared memory and using queues of different sizes to transfer experience. 
It can be seen from Fig.~\ref{fig_aba}(a) that the training effect of the queue transmission method is sensitive to the queue size.
Even if the queue size is fine-tuned, the queue method is still inferior to the shared memory method because the use of shared memory can greatly reduce the experience transmission cycle without occupying the network update process time.
The quantitative results of the throughput using queues to transfer experience are shown in rows 6$\sim$8 of Table~\ref{tab_para}. 

On the other hand, we conduct hardware device limitation experiments by restricting the framework's use of CPU and GPU devices.
Fig.~\ref{fig_aba}(b) shows the experiments of the Spreeze framework using different amounts of CPU resources, namely 100\% CPU, 50\% CPU and 25\% CPU.
As the use of CPU hardware resources is restricted, the training effect has slightly decreased. This shows that it is meaningful for our framework to use a large number of experience sampling processes to sample experiences in parallel and at high speed.
Similarly, Fig.~\ref{fig_aba}(c) shows the results of the experiment that restricts the use of the GPU.  
By default, the framework uses two GPUs for model parallel network update.
In Fig. 5(c), GPU1 indicates that only one graphics card is used for network training without model parallelization, which will cause the network update throughput to drop by about 10\%, and the final training curve will be slightly affected.
Furthermore, restricting the framework to only use 75\% or 50\% of a single GPU will further significantly deteriorate the training effect.
Compared to restricting the use of the CPU, restricting the GPU will have a greater impact on training performance. This also shows that the focus of the future performance improvement direction of the parallel framework is to improve the efficiency of network updates under the conditions of large batch training.

\begin{figure}[t]
	\centering
	\subfigure[Batch Size]{
		\includegraphics[width=0.65\linewidth]{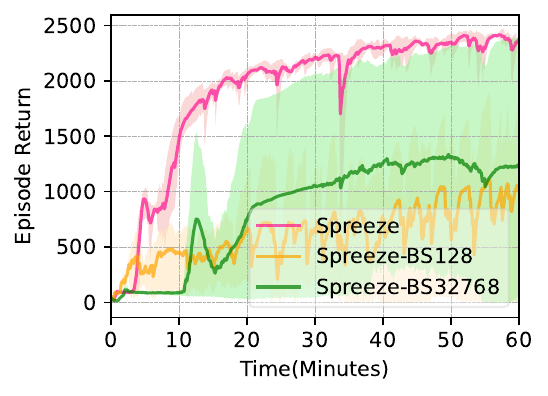}
	}
	\subfigure[Number of Sample Processes]{
		\includegraphics[width=0.65\linewidth]{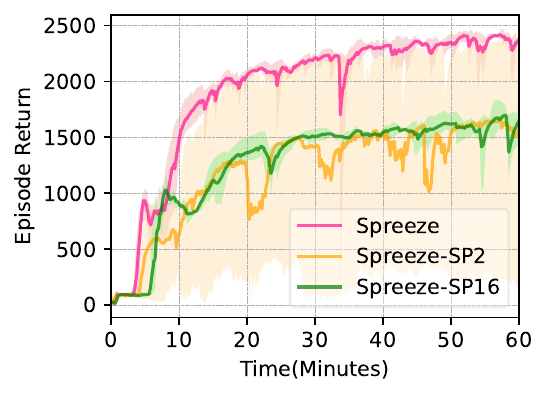}
	}
	\caption[]{\textbf{The effect of hyperparameters on the final training performance of our framework.} Each curve represents the average of five random seed experiments.}  
	\label{fig_para}
\end{figure}

Fig.~\ref{fig_para} shows the training curve for adjusting the hyperparameters of the batch size and the number of sampling processes. The performance of the adjusted hyperparameters is not as good as the performance with the 8192 batch size and 16 sample processes automatically determined by the framework.

The first row of experimental data in the Table~\ref{tab_para} represents the result of the framework automatically determining the hyperparameters. The batch size is automatically determined to be about 8192, and the number of sampling processes is determined to be about 16. The 2$\sim$5 rows in the table are the results obtained after manually changing one parameter.
As shown in Fig.~\ref{fig_para}, the performance after manually changing the parameters is not as good as that of the default parameters automatically determined by the framework. 

\begin{figure}[tb]
	\centering
	\subfigure[Device robustness.]{
        \includegraphics[width=0.55\linewidth]{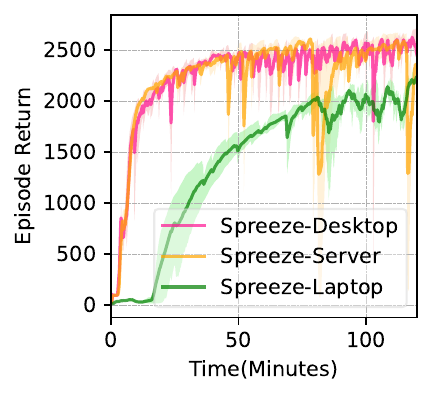}
    }
    \subfigure[Algorithm robustness.]{
        \includegraphics[width=0.55\linewidth]{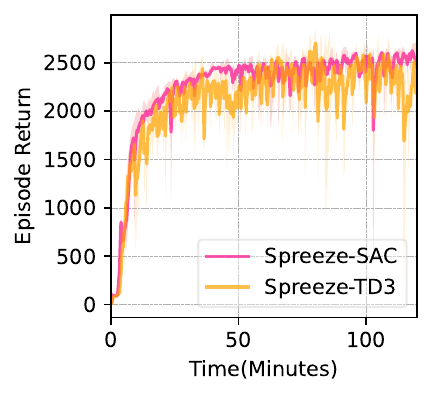}
    }
    
	\caption[]{\textbf{Robustness experiment.}
	Five random seed trainings are performed under each experimental condition.}
	\label{fig_device}
\end{figure}

In addition, it can be seen from Table~\ref{tab_para} that the batch size parameter mainly affects network updates and GPU usage.
Compared with the default setting of 8192, when the batch size is set too large to 32768, although the network update frame rate can be slightly increased, the network update frequency is significantly reduced, thereby deteriorating the training effect.
When the batch size is as small as 128, the network update frame rate will be reduced by an order of magnitude, and GPU utilization will also be significantly reduced.
The number of sampling processes mainly affects experience collection and CPU usage. When it is too high, the sampling frame rate and CPU usage are increased, but the experience transmission loss is aggravated, which in turn takes up the network update time.
When it is too low, the training effect is poor because the experience sampling frame rate is low and the collected experience is not rich enough.

And if the program uses queue transfer experience instead of memory sharing experience, the size of the queue will also be a key parameter that affects parallelization performance. A large queue size can reduce the time occupied by the experience transmission in the network update process and improve the experience transmission efficiency. But this will also make the experience transmission cycle longer, and the network update process will not get the data sampled with the latest strategy.

\subsubsection{Robustness experiment}
As a general RL framework, our Spreeze framework can be used on a variety of hardware devices and algorithms. We have conducted experiments to ensure that our framework can maintain the best possible training effects on a variety of hardware devices and algorithms.
In addition to the desktop computer used in the previous experiments, a computing server with 40-core Intel 5128R CPU and Nvidia 2080Ti GPU, and a laptop with 4-core Intel 4710MQ CPU and Nvidia 950M GPU are used for the robustness experiment. The experiment is trained in the PyBullet walker2D environment for 120 minutes to evaluate the reward curve.
As described in section 3.3, our framework automatically adjusts the parallelization hyperparameters for different hardware devices. On the server, the batch size and the number of sampling processes are set to 16384 and 16 respectively, while on the laptop, these two parameters are set to 2048 and 4.
The result of the training curve is shown in Fig.\ref{fig_device}(a). 
The server's GPU is not much better than that of the desktop, so the server robustly maintains a training effect similar to that of the desktop. For the laptop, due to its poor GPU computing power, its training curve is significantly worse than that of the desktop computer and the server.
In general, as much parallelization as possible has been achieved on each computing device. The training effect that is proportional to the computing power of the hardware device verifies that our framework has good robustness and gives full play to the capabilities of each device.

In addition to the SAC algorithm, our framework can also be easily extended to off-policy RL algorithms such as TD3 \cite{fujimoto2018addressing}. The experimental results of the robustness of different algorithms are shown in Fig.\ref{fig_device}(b), and each algorithm can be well parallelized. As a result, under the strong parallelization, the performance gap between the algorithms appears to be quite small.

\section{Conclusion}

\label{conclusion}

In summary, this paper proposes a parallel RL framework that fully exploits hardware devices, and achieves high throughput via efficient data transmission, parameter adaptation, and network update, and reduces training time by an average of 73\%.
Our framework is compatible with the widely used OpenAI-gym system environment and significantly accelerates RL training on the most prevalent single desktop platform.

Making full use of hardware devices on a single desktop is the basis for realizing a high-throughput distributed parallel training framework.
Our framework mainly uses parallelization to more reasonably arrange computing tasks on different computing hardware devices, and does not optimize the internal algorithm principles of RL. So the calculation time is reduced affinely so that the calculation time complexity is still $O(S^2 A)$. We look forward to further breakthroughs in parallel computing theory in the future.
Our framework is dedicated to advancing RL into an era of more extensive, practical, user-friendly, and efficient large-scale applications.

\ifCLASSOPTIONcompsoc
  \section*{Acknowledgments}
\else
  \section*{Acknowledgment}
\fi

This work is funded by the Shanghai Municipal Science and Technology Major Project (No.2018SHZDZX01) of ZJ Lab and Shanghai Center for Brain Science and Brain-Inspired Technology, Shanghai Rising Star Program (No. 21QC1400900) and Tongji-Westwell Autonomous Vehicle Joint Lab Project.

\ifCLASSOPTIONcaptionsoff
  \newpage
\fi

\bibliographystyle{ieeetr}
\bibliography{ref}

\end{document}